# Rethinking Reprojection: Closing the Loop for Pose-aware Shape Reconstruction from a Single Image

Rui Zhu, Hamed Kiani Galoogahi, Chaoyang Wang, Simon Lucey
The Robotics Institute, Carnegie Mellon University
{rz1, hamedk, chaoyanw}@andrew.cmu.edu, slucey@cs.cmu.edu

## Abstract

*An emerging problem in computer vision is the reconstruction of 3D shape and pose of an object from a single image. Hitherto, the problem has been addressed through the application of canonical deep learning methods to regress from the image directly to the 3D shape and pose labels. These approaches, however, are problematic from two perspectives. First, they are minimizing the error between 3D shapes and pose labels - with little thought about the nature of this "label error" when reprojecting the shape back onto the image. Second, they rely on the onerous and ill-posed task of hand labeling natural images with respect to 3D shape and pose. In this paper we define the new task of pose-aware shape reconstruction from a single image, and we advocate that cheaper 2D annotations of objects silhouettes in natural images can be utilized. We design architectures of pose-aware shape reconstruction which reproject the predicted shape back on to the image using the predicted pose. Our evaluation on several object categories demonstrates the superiority of our method for predicting pose-aware 3D shapes from natural images.*

## 1. Introduction

Reliably predicting the 3D shape and pose of an object from a single image has only become feasible in computer vision over the last few years due to advances in deep learning. A substantial barrier to success stems from the inherent lack of natural training images with labeled 3D shape and pose information. Some efforts have been undertaken recently to rectify this including the construction of PASCAL3D+ [27], ObjectNet3D [26] and IKEA [13] datasets. In all these cases images of several object categories are manually annotated with corresponding 3D CAD models and pose information.

Such datasets, however, suffer from several limitations. First, they are limited to very few object categories and samples. Second, the human labeler must choose a 3D CAD model from a finite dictionary of models. This is problematic and error prone as the dictionary of CAD models made available rarely covers the actual variation encountered in natural imagery (e.g. only 7 CAD models are used to describe all natural images of category "aeroplane" in PASCAL3D+). Third, the pose of the CAD model is determined by annotating coarse landmarks within the natural image that correspond to points on the CAD model - pose is then recovered through a PnP process [15]. This is also error prone due to the shape mismatch of the CAD model with the natural image. Given the above limitations, directly training or even fine-tuning with such datasets is not desirable.

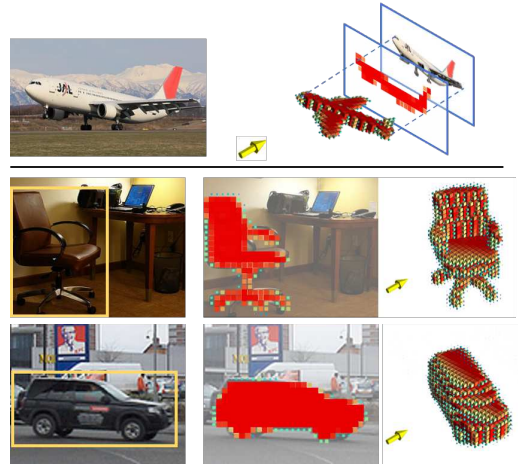

Figure 1: Input natural images with bounding box (left), reprojected silhouettes (middle), and reconstruction (right) results. The yellow arrow shows the canonical camera.

Recently, the vision community has shifted their attention to synthetically rendering images directly from textured CAD models [19, 2]. Notable efforts that use synthetic imagery for training deep models to estimate pose or reconstruct 3D shape include [19, 6, 1, 4, 11, 17, 20, 18]. This trend offers two advantages. First, it alleviates the need for using images with error prone hand labeled pose and 3D shapes. Further, the textured CAD models can be used to synthesize nearly limitless amounts of realistic rendered training images with accurate ground truth pose and 3D model labels [19]. Second, the proven potential of deep networks to model complicated patterns can be exploited to handle large amounts of appearance variations [6, 3, 24]. While these approaches have shown promising results on rendered images, there is a noticeable drop in performance [17] when applied to natural (i.e. non-rendered) images. We shall refer to this herein as the "render gap". Efforts have been made to fine-tune these networks using a small amount of labeled natural images [3] to overcome this problem but the intrinsic errors associated with the labeling process limits its effectiveness.

Unlike labelling object shape and pose, annotating the silhouette of an object within a natural image can be performed extremely efficiently and accurately by human labelers. Instance segmentation tools (as they are often referred to) have evolved in such a manner that are has now become feasible to hand segment tens of thousands of images from an object category for a reasonable amount of cost and effort (e.g. more than 13k segmentation for chairs in MS COCO [14]).

**Contributions:** In this paper we want to explore how one could take advantage of this, hitherto, untapped resource for training a deep network for predicting 3D shape and pose that addresses the "render gap". Specifically, we propose a novel scheme to extend current state-of-the-art methods [24, 6] to predict *pose-aware* 3D (voxelized) shape of an object from a single natural image using cheap silhouette labels of natural images (see Fig. 1). The key observation behind our approach is, the pose-aware shape estimate should match well with the silhouette of the object in the image when reprojected back to the image plane. This insight allows us to leverage a much larger set of hand segmented natural images for training a deep network that alleviates the "render gap".

Our method differs from previous related works [29, 6, 24, 3] in several ways. First, our method is capable of learning from both rendered image-shape pairs as well as natural images with annotated silhouettes - the only approach thus far to our knowledge to do so. Second, unlike [6, 29, 24, 3] which output 3D voxelized shapes in canonical viewpoint (we refer to as *aligned shapes* throughout the paper), our approach also simultaneously predicts the shape in full 6 DOF of pose. During training we jointly optimize over pose and style as our proposed reprojection metric is dependent on both. This differs from previous approaches that learn pose [19] and style (i.e. aligned shapes) [6, 24, 3] independently. Third, we argue that reprojection error rather than 3D reconstruction error is a preferable loss for training deep networks for predicting pose and 3D shape when single natural images lack 3D shape ground-truth.

Our proposed scheme is applicable to current approaches of predicting aligned shapes, such as 3D-VAE-GAN [24] or the TL-embedding network [6]. As illustrated in Fig. 2, we build the architecture of pose-aware reconstruction based on these two methods (see Appendix I in supplementary material), respectively named as **p-TL** and **p-3D-VAE-GAN**. Fine-tuning on natural images is performed in our training pipeline with a novel reprojection loss. At testing time, an input natural image of an object is fed into the fine-tuned network (Fig. 2(part 3)) to estimate its pose-aware 3D shape. No silhouette is needed in the testing stage. More details are provided throughout the rest of the paper.

This strategy closes the loop for pose-aware shape reconstruction both in training and testing: being able to reproject back onto the image frame gives an extra metric of how nice the reconstruction is by just looking at how well the reprojected shape matches the object in the image. We also demonstrate later that, in our scheme this metric of reprojection error also speaks for quality 3D reconstruction; in other words, out strategy does not degrade 3D reconstruction performance.

**Notation:** Vectors are represented with lower-case bold font (*e.g.* **a**). Matrices are in upper-case bold (*e.g.* **M**) while scalars are italicized (*e.g. a* or *A*). Variables with a subscript $gt$, *e.g.* $\mathbf{M}_{gt}$, are the ground truth of the corresponding variable **M**. For denoting the $l^{th}$ sample in a set (e.g. images, shapes), we use superscript with parentheses (*e.g.* $\mathbf{M}^{(l)}$). Uppercase calligraphic symbols (e.g.$\mathcal{F}(\mathbf{x})$) denote functions which take in a vector or a scalar.

## 2. Related Work

**Shape from a Single Image: Approaches and Data** Non-learning approaches address the problem of shape reconstruction from a single image mainly through optimization. Shape priors are either acquired from CAD datasets [25], or learned with structure from (sub)category techniques [22, 9]. In these methods weaker annotations are often required for optimizing shape or pose, such as key points [22] or instance segmentation [9], limiting the application and performance of these methods. Moreover, imperfect shape prior and optimization procedure have led to smooth but fuzzy reconstructions, and pose optimization from a single image has been error-prone.

For learning based methods, Xiang et al. [25] develop exemplar detectors of pose-aware shapes through annotated image-shape pairs. Such detectors, however, are acquired with very limited amount of natural images. This drastically limits the application of this method for 3D inference in the *wild*, where objects display severe occlusion, uncommon pose, and large intra-class variation. To alleviate the problem, domain adaptation has been applied in the feature level with rendered-natural image pairs [17]. This methods

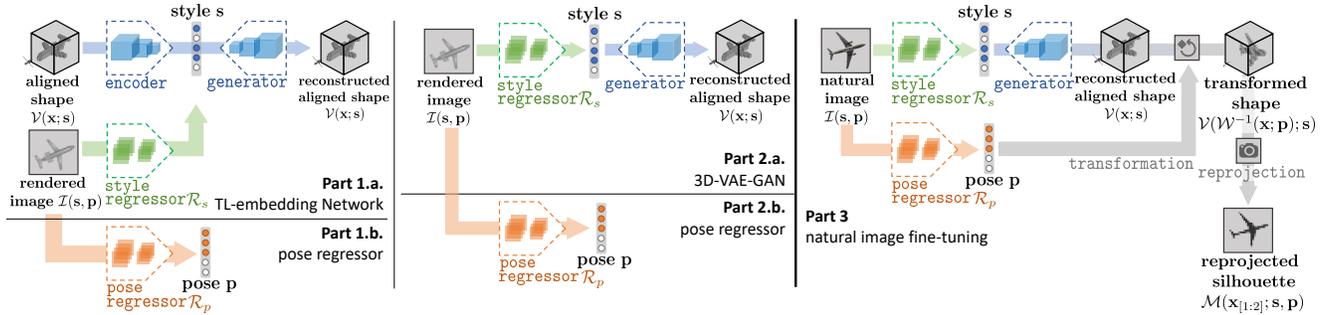

Figure 2: The proposed methods for reconstructing pose-aware 3D voxelized shapes (details in the text): **p-TL** (part 1 & 3) and **p-3D-VAE-GAN** (part 2 & 3).

still requires annotated natural image-shape pairs.

Recently the emerging field of deep 3D vision has witnessed rapid development in this task. Among these methods, the TL-embedding network [6], 3D-VAE-GAN [24] and 3D-R2N2 [3] take advantage of generative networks to embed 3D representation in latent space as shape prior, and develop regressors from image domain to the shape domain. Moreover, 3D-R2N2 introduces LSTM to the hidden representation to accommodate sequential inputs. The main drawback of these methods, however, is that they are either purely trained on rendered samples [6, 24], or fine-tuned on very few natural image-shape pairs [3]. This nature unavoidably limits the generalization ability of these methods to natural testing images due to the statistical difference between features extracted from rendered images and the natural images ("render gap").

**Representation and Factors of Shape** In the age of non-learning methods, mesh representation [9, 22] is prominent due to its flexibility during optimization. In later learning-based methods, voxelized shapes are in more favor because the quantized voxel grid they live in enables easy labeling [25] and better suits convolutional operations in deep learning [24, 6, 3]. Skeleton (connections among key points) has also been considered [23] but beyond the discussion of shape in this paper.

For the factors influencing shapes, most works consider shapes in canonical viewpoint (pose). In this case, only style variations are parametrized. However, datasets have been developed for pose-aware shape annotations, e.g. PASCAL3D+ [27] and IKEA dataset [13]. A few works take advantage of these datasets and perform pose-aware reconstruction [9, 22, 23]. These works mostly infer the correct pose and shape in mesh form in a non-learning manner, and require instance segmentation as input [9, 22]. Another stream of works focus on estimating pose alone from a single image, e.g. [21, 17] and [19].

**Reprojection Loss as Supervision** The role of reprojection loss in related tasks has been explored in various occasions, including constraining 3D reconstruction with multi-view annotations [29], weakly-supervised shape optimization [9, 22], or structure from silhouettes [5]. However in the task of learning shape from natural images, the rich repository of segmentation annotations on natural image sets have never been explored as supervision.

## 3. Proposed Method

### 3.1. Assumption

We follow [6, 24, 3] by making the following assumptions. (1) We assume weak-perspective projection to avoid estimating camera intrinsics. (2) We have the bounding box of the object throughout the approach. (3) Shape is represented as binary voxels in a regular grid, written as a function $\mathcal{V}(\mathbf{x}) = \{0, 1\} : \mathbb{R}^3 \to \mathbb{B}$, which is sampling the single channel voxelized shape at location $\mathbf{x} = [x, y, z]^T$ in the voxel grid. (4) Realistic rendering is done with random lighting and original texture from the CAD models as an approximation to real-world statistics.

### 3.2. Shape Parametrization

We propose to model 3D pose-aware shape of an object by both *pose* and *style* parameters. We can express a shape as a function $\mathcal{V}(\mathbf{x}; \mathbf{s}, \mathbf{p})$ parametrized by the style parameter $\mathbf{s} \in \mathbb{R}^M$ and pose parameter $\mathbf{p} \in \mathbb{R}^N$. These parameters will be recovered at testing time, and used to estimate pose-aware 3D shapes from natural images. By this definition, shapes in canonical pose but varying in styles are named "aligned shapes".

In the recent works of [6, 24, 3] only shapes in canonical pose are considered, reflecting variations in style. In these works, a generative network is applied to learn an embedding of style parameters and a generator to recover the shape from the embedding. However, for pose-aware shapes, they are also heavily influenced by the 6 DOF pose of the object, or equivalently, the extrinsics of the camera. Hence, pose-aware shapes can be readily embedded in a space with style and pose parameters coupled together. However, we argue that this is a suboptimal goal to aim for because the space of pose-aware shapes grows in a multiplicative manner with the degree of freedom imposed by the pose parametrization, demanding accordingly increased capacity of the generative model.

Instead, given 6 DOF pose can be *explicitly* parametrized with $\mathbf{p} \in \mathbb{R}^N$ in our camera model, we propose to decouple style and pose parameters by estimating explicit pose independently from style, and apply an rigid transformation operation to the aligned shape generated from style parameters to impose pose. This results in unchanged capacity demand of the generative model, and an additively increased parameter space rather than multiplicatively. This discussion echoes with [11] where transformations of head pose in a facial image are disentangled from appearance variations when learning an autoencoder. Experiments can be found in Section 4.2 where a comparison between learning with coupled and decoupled style and pose parameters is drawn in the case of 3D VAE [10] and 3D-VAE-GAN [24].

### 3.3. Overview of the Training Pipeline

The approach consists of three main stages. (1) Train an style regressor and a shape generator which maps from a rendered image to an aligned shape. (2) Train a pose regressor which regresses from the image to the pose parameters. Note that this training procedure can be carried out in parallel with the first step. (3) Append a pose transformation and reprojection layer after the generator to transform the aligned shape according to the estimated pose and then reproject it back to the image frame. Fine-tune the style regressor and pose regressor to natural image sets by minimizing reprojection loss between the ground truth silhouettes and reprojected ones. We build two architectures for pose-aware shape reconstruction following our approach, under the name of **p-TL** (pose-aware TL-embedding network; part 1 & 3 in Fig. 2) and **p-3D-VAE-GAN** (pose-aware 3D-VAE-GAN; part 2 & 3 in Fig. 2), respectively upon TL-embedding network [6] and the state-of-the-art 3D-VAE-GAN [24].

**Encoder and Generator for Aligned Shapes:** For the first stage, the architecture is explored in works of [6, 24, 3]. We build our architectures of style encoder and aligned shape generator upon [6] and [24] with minor improvements, and leave [3] for future exploration.

In our version of **p-TL** (see part 1.a in Fig. 2), the vanilla autoencoder is replaced with more recent volumetric variational autoencoder (VAE) [10] to learn a compact style embedding space for aligned shapes. After the VAE is learned, a style regressor connects images to style space. For **p-3D-VAE-GAN**, we adapt the architecture from [24] (see part 2.a in Fig. 2) to fit the grid size of 30×30×30, and use reconstruction loss of convolution features instead of voxels as suggested in [12]. Network parameters and training details can be found in Appendix I in supplementary material.

**Image to Pose Regressor:** At the second stage, we train an extra pose regressor (part 1.b & 2.b in Fig. 2) to regress rendered images to their ground truth pose parameters. Given 3-channel rendered RGB images as a function of subpixel location $\mathbf{u} = [u, v]^T$, parametrized by the pose parameters $\mathbf{p}$ and style parameters $\mathbf{s}$ of the object in the image: $\{\mathcal{I}^{(l)}(\mathbf{u}; \mathbf{s}^{(l)}, \mathbf{p}^{(l)}) : \mathbb{R}^2 \to \mathbb{R}^3\}_{l=1}^L$, we train a pose regressor $\mathcal{R}_p(\mathcal{I}(\mathbf{s}, \mathbf{p})) = \mathbf{p}$ to map an image $\mathcal{I}$ to ground truth pose $\mathbf{p}_{gt}$ by minimizing the Euclidean loss between $\mathbf{p}$ and $\mathbf{p}_{gt}$. Here we abuse the notation a bit by denoting $\mathcal{I}(\mathbf{s}, \mathbf{p})$ as a concatenation of $\mathcal{I}(\mathbf{u}; \mathbf{s}, \mathbf{p})$ over all pixels:

$$\mathcal{I}(\mathbf{s}, \mathbf{p}) = \begin{bmatrix} \mathcal{I}(\mathbf{u}_1; \mathbf{s}, \mathbf{p}) \\ \vdots \\ \mathcal{I}(\mathbf{u}_D; \mathbf{s}, \mathbf{p}) \end{bmatrix} \in \mathbb{R}^{3D \times 1} \qquad (1)$$

**Pose-aware Shape Reconstruction and Fine-tuning on Natural Images:** The last part of our framework (Fig. 2 part 3) fine-tunes style and pose regressors on natural images. We denote the style regressor as $\mathcal{R}_s(\mathcal{I}(\mathbf{s}, \mathbf{p})) = \mathbf{s}$. The aligned shape generator takes in $\mathbf{s}$ and outputs the reconstructed aligned shape $\mathcal{V}(\mathbf{x}; \mathbf{s})$.

To impose the correct pose on the reconstructed shape, we design a rigid transformation layer to transform the reconstructed aligned shape with the predicted pose of $\mathbf{p}$. We define the transformation function as an inverse warp parametrized by $\mathbf{p}$: $\mathcal{W}^{-1}(\mathbf{x}; \mathbf{p}) : \mathbb{R}^3 \to \mathbb{R}^3$, which will be discussed later. In such case, the transformed shape in vectorized form is

$$\mathcal{V}(\mathbf{s}, \mathbf{p}) = \begin{bmatrix} \mathcal{V}(\mathcal{W}^{-1}(\mathbf{x}_1; \mathbf{p}); \mathbf{s}) \\ \vdots \\ \mathcal{V}(\mathcal{W}^{-1}(\mathbf{x}_{Q^3}; \mathbf{p}); \mathbf{s}) \end{bmatrix} \in \mathbb{R}^{Q^3} \qquad (2)$$

where $Q$ is the side length of the voxel grid. The inverse warp can be implemented as inverse sampling of original voxels with the rigid transformation parametrized by $\mathbf{p}$. This transformation layer retains the style of the predicted shape and only changes its pose.

A reprojection operation is then applied after the transformation layer to project the rotated voxels onto the image plane, assuming that a weak-perspective projection camera is placed at a fixed canonical viewpoint. This returns a mask $\mathcal{M}(\mathbf{x}_{[1:2]}) \in [0, 1] : \mathbb{R}^2 \to \mathbb{R}$ where $\mathbf{x}_{[1:2]} = [x, y]^T$. Assuming the shape $\mathcal{V}(\mathcal{W}^{-1}(\mathbf{x}; \mathbf{p}); \mathbf{s})$ is reprojected along its $3^{rd}$ dimension (which we assume as axis $z$ here), then we write the reprojection as:

$$\mathcal{M}(\mathbf{x}_{[1:2]}; \mathbf{s}, \mathbf{p}) = \max_z \mathcal{V}(\mathcal{W}^{-1}(\mathbf{x}; \mathbf{p}); \mathbf{s}) \qquad (3)$$

The max operation can be considered as performed along the optical axis of the camera, analogous to ray-tracing where the value of first non-zero cubic is returned.

Given a ground truth natural image with annotated silhouette $\mathcal{M}_{gt}(\mathbf{x}_{[1:2]}; \mathbf{s}, \mathbf{p})$ and the predicted silhouette $\mathcal{M}(\mathbf{x}_{[1:2]}; \mathbf{s}, \mathbf{p})$, the reprojection loss is defined with binary cross-entropy:

$$\mathcal{L}_{rp}(\mathcal{M}, \mathcal{M}_{gt}) = \qquad (4)$$
$$\frac{1}{Q^2} \sum_{j=1}^{Q^2} -\mathcal{M}_{gt}(\mathbf{x}_{[1:2]j}; \mathbf{s}, \mathbf{p}) \log(\mathcal{M}(\mathbf{x}_{[1:2]j}; \mathbf{s}, \mathbf{p}))$$
$$- (1 - \mathcal{M}_{gt}(\mathbf{x}_{[1:2]j}; \mathbf{s}, \mathbf{p})) \log(1 - \mathcal{M}(\mathbf{x}_{[1:2]j}; \mathbf{s}, \mathbf{p}))$$

The network is then fine-tuned end-to-end by minimizing reprojection loss on natural images. The shape generator is fixed in this step to function as a prior on the learned style space. Moreover, natural images are mixed with rendered images in equal portions within a batch to stabilize fine-tuning. Details of fine-tuning will be shown later.

**Pose Parametrization:** In our setting, we intend to recover full 6 DOF pose, which is 3 DOF of rotation and 3 DOF of translation in a weak-perspective model. However we find the generative model learns well the interpolation between shapes of varying scale, thus we treat the DOF of scale (depth) as one latent factor of style, that is, an aligned shape under different scales are considered to be different in style, instead of in pose. The repository of CAD models is then accordingly augmented with random scale. As a result our model recovers shapes in full 6 DOF by using pose parameters $\mathbf{p} \in \mathbb{R}^5$ (3 for rotation plus 2 for in-plane translation).

For rotation, the parametrization of Euler angles [19] works well in pose classification tasks among limited viewpoints. Euler angles, however, suffer from the issue of gimbal lock [7][1] and non-uniformly distributed pose space. Another popular choice for rotation parametrization is quaternion. This parametrization, however, is strongly limited by the unit norm constraint which is not suitable for regression tasks. We choose to parametrize rotation using the exponential twist, where an rotation around an unit axis for less or equal than 180° clockwise is parametrized by the axis vector and rotation angle. More specifically, for a rotation around unit axis $\mathbf{n} = [n_1, n_2, n_3]^T$ for radian $\phi$, the rotation matrix is given by $\mathbf{R} = e^{[\mathbf{n}]_\times \phi}$, where $[\mathbf{n}]_\times$ is the skew-symmetric matrix, defined as:

$$[\mathbf{n}]_\times = \begin{bmatrix} 0 & -n_3 & n_2 \\ n_3 & 0 & -n_1 \\ -n_2 & n_1 & 0 \end{bmatrix} \qquad (5)$$

By setting the rules for exponential twist, the twist parameters $\mathbf{w} = \phi \mathbf{n} \in \mathbb{R}^3$ is only constrained within a ball of radius $\pi$ in $\mathbb{R}^3$. This constraint can be implemented with the *tanh* function to limit the norm of $\mathbf{w}$. Another interesting property of exponential twist is that the derivative of $\mathbf{R}$ with respect to $\mathbf{w}$ is readily available which makes it feasible for optimization with first-order methods [8]. By explicitly parametrizing pose with exponential twist, the dimension of rotation parameters is 3. For in-plane translation, two scalars are represented in $\mathbf{t} = [t_x \; t_y \; 0]^T \in \mathbb{R}^3$. Thus the final inverse warp function can be written as:

$$\mathcal{W}^{-1}(\tilde{\mathbf{x}}; \mathbf{p}) = \begin{bmatrix} \mathbf{R} & \mathbf{t} \\ \mathbf{0} & 1 \end{bmatrix}^{-1} \tilde{\mathbf{x}} \qquad (6)$$

where $\tilde{\mathbf{x}}$ is the homogeneous coordinates of $\mathbf{x}$ in the voxel grid, $\mathbf{p} = [\mathbf{w}^T \; t_x \; t_y]^T \in \mathbb{R}^5$.

---
[1] when elevation of the camera reaches 90°, 1 degree of freedom is lost between azimuth and yaw.

As for the choice between classification and regression for pose estimation as discussed in [16], we favor regression over classification because, in our case the estimated pose needs to be converted to a transformation matrix as in Eq. 6, and this conversion as a function needs to be smooth in the step of fine-tuning. However this would not be the case if we follow [19] by outputting an one-hot vector of $\mathbf{p}$ over the pose space, considering we will need a lookup table for this purpose, which in nature is not differentiable.

## 4. Experiments

### 4.1. Data Preparation

Given there is no publicly available dataset for our purpose, we collect and process the data from some existing datasets. We select three object categories, including aeroplane, chair, and car which have datasets available for both 3D CAD models (for rendering training data), natural images with annotated segmentation masks (silhouettes for natural image fine-tuning) and natural images with annotated 3D shapes (for evaluation). We use CAD models from the ShapeNet dataset [19] and their CAD-to-voxel pipeline to voxelize CADs into $30 \times 30 \times 30$ voxel grids. We use the rendering procedure provided in [19] to generate rendered images with sampled lighting and 6 DOF poses, over background natural images from the SUN dataset [28]. Poses are sampled from the distribution in PASCAL3D+ dataset [27] with random perturbations. The natural image-silhouette pairs for fine-tuning are obtained from the instance segmentation masks in the MS COCO [14] dataset, cropped, normalized and centered in the $227 \times 227$ image frame. We pruned the dataset beforehand to remove samples with strong perspective effect or severe occlusion to facilitate natural image fine-tuning as we did not take care of such situations in our assumptions. Natural image-shape pairs for testing are acquired from PASCAL3D+ with ground truth pose and style annotation, and the same preprocessing is applied as to the data from MS COCO. The size of data we used for training and testing is listed in Table. 1.

|  | aeroplane | chair | car |
|---|---|---|---|
| rendered with shapes | 206,296 | 345,001 | 382,144 |
| MS COCO with masks | 4,734 | 3,200 | 2,942 |
| PASCAL3D+ with shapes | 125 | 220 | 279 |

Table 1: Size of data for each object category.

|  | aligned shape | | pose-aware shape | |
|---|---|---|---|---|
|  | V2V | im2V | V2V | im2V |
| 3D VAE | 0.876 | - | 0.544 | - |
| 3D-VAE-GAN | - | 0.752 | - | 0.403 |
| p-3D-VAE-GAN | - | - | - | 0.665 |

Table 2: Average precision (AP) of learning 3D VAE, 3D-VAE-GAN and p-3D-VAE-GAN with aligned shapes and pose-aware shapes of three categories (car, aeroplane, and chair).

### 4.2. Pose as a Latent Shape Factor

As discussed in Section 3.2, decoupling pose from style when parametrizing pose-aware shapes helps to reduce the demand for model capacity, hence improves reconstruction performance. To evaluate this assertion, we develop two experiments on rendered dataset, including (1) volumetric reconstruction with 3D VAE, and (2) volumetric reconstruction from single image with 3D-VAE-GAN and p-3D-VAE-GAN using rendered image-shape pairs for both aligned and pose-aware shapes. The results can be found at Table 2 measuring average precision (AP) of reconstruction averaged over three categories, following the practice of [24]. In this table, V2V shows the reconstruction AP of the output voxels compared to input voxels, and im2V shows the rfeconstruction AP of the output voxels to the ground truth voxel of the input image. See Appendix I in supplementary material for details of the implementation.

The first experiment of 3D VAE shows drastically dropped performance when trying to encode pose-aware shapes compared with encoding aligned models (V2V aligned versus V2V pose-aware shape reconstruction), since the space of pose-aware shapes is substantially larger than that of aligned shapes. Thus, generative models such as VAE are not able to efficiently encode such a large space of style parameters. The second experiment shows that compared to 3D-VAE-GAN, p-3D-VAE-GAN performs much better at learning generative models to directly generate pose-aware shapes from natural images (im2V pose-aware shape reconstruction). This demonstrates the efficiency of p-3D-VAE-GAN which decouples style from pose over two separate regressors for pose-aware shape reconstruction, instead of treating pose as part of the latent parameters and learning a latent representation of style and pose altogether.

### 4.3. Qualitative Evaluation

We evaluate the performance of our approach on PASCAL3D+ images with annotated masks, pose and 3D shapes, before and after fine-tuning. However, it should be noted that although we use these annotations as approximate ground truth in our setting of low resolution shapes, the limitations of this dataset- as we mentioned earlier- still make it a suboptimal target to evaluate against.

Fig. 3 and Fig. 4 visualize qualitative results showing style and pose improvements after fine-tuning on natural images, respectively. The yellow arrows show the canonical camera, and larger cubes with warmer colors indicate higher confidence scores. At Fig. 3, we show ground truth and reconstructed shapes (before and after fine-tuning) in both canonical and predicted poses (*i.e.* pose-aware shape reconstruction). The effect of fine-tuning on predicting more accurate styles can be seen by comparing the reconstructed aligned shapes and reprojected silhouettes before and after fine-tuning against ground-truth. Fig. 4 shows the improvement in pose estimation after fine-tuning. For this qualitative evaluation, we show pose-aware ground truth as well as pose-aware shapes predicted before and after fine-tuning. Again, by comparing the ground truth silhouettes with those generated before and after fine-tuning, one can see that silhouettes after fine-tuning are visually more similar to ground truth. These qualitative results show that fine-tuning improves the estimated pose and projected silhouettes. Moreover, these qualitative results demonstrate that in addition to predicting more accurate pose and style, natural image based fine-tuning improves the robustness against ambiguous poses. For the case of the first sample of aeroplane in Fig. 4 (top left), the shape recovered before fine-tuning suffers from ambiguity in azimuth. By constraining the reprojected aeroplane towards its ground truth body length, the pose parameters are optimized in the continuous pose space (particularly the DOF of azimuth in this example) in a way to produce more accurate silhouette. This significantly diminishes the pose ambiguity.

**Failure Cases Analysis:** Fig. 5 illustrates some failure cases, which either end up with wrong shape before fine-tuning, or fail to improve after fine-tuning. We categorize the failure cases generally into three scenarios, including (1) Ambiguous pose: Fig. 5 (up), (2) Occlusion: Fig. 5 (middle), this sample was mistakenly labelled clean in PASCAL3D+ but in fact seriously occluded, leading to broken results, and (3) Low image quality: Fig. 5 (bottom). Deteriorated (murky) images may result in wrong reconstruction because their appearance statistics deviate too far from that of the photorealistically rendered training images.

### 4.4. Quantitative Evaluation

In Table 3 and 4, we evaluate our performance by means of AP between 3D shapes (3D AP) or 2D silhouettes (2D

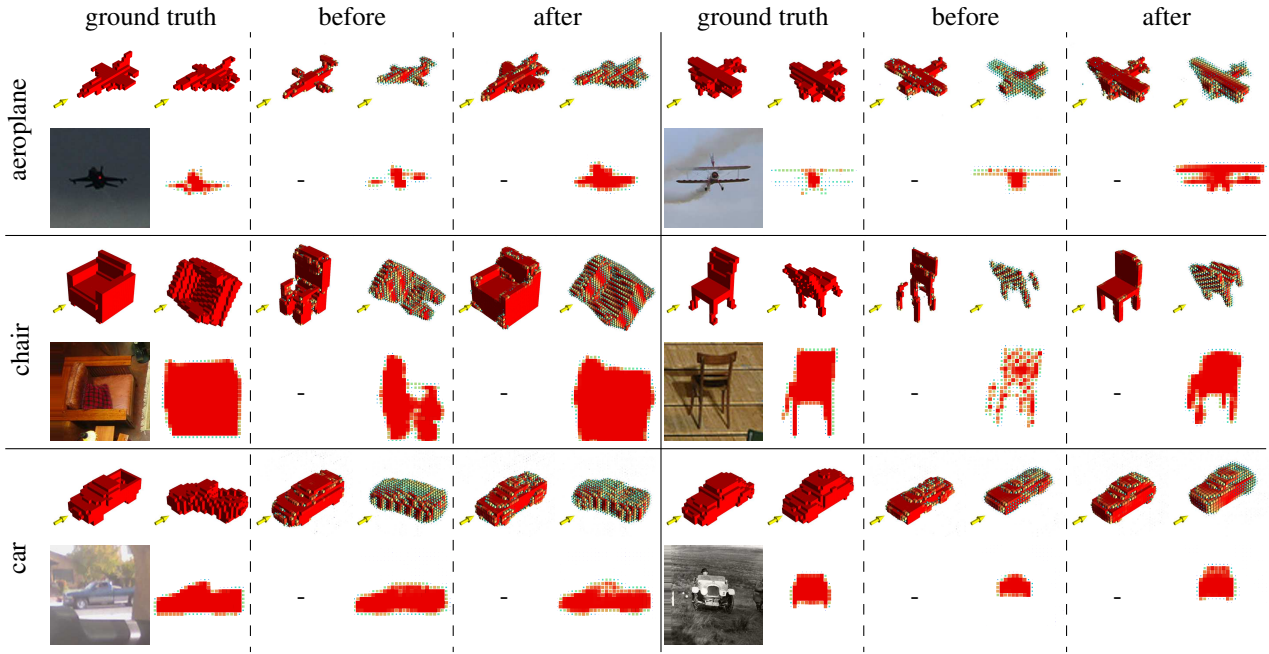

Figure 3: **Examples of improvement in style with fine-tuning.** For each sample, illustrations include: input image, aligned & shape-aware shapes, reprojected silhouette of ground truth (left), before fine-tuning (middle) and after fine-tuning (right).

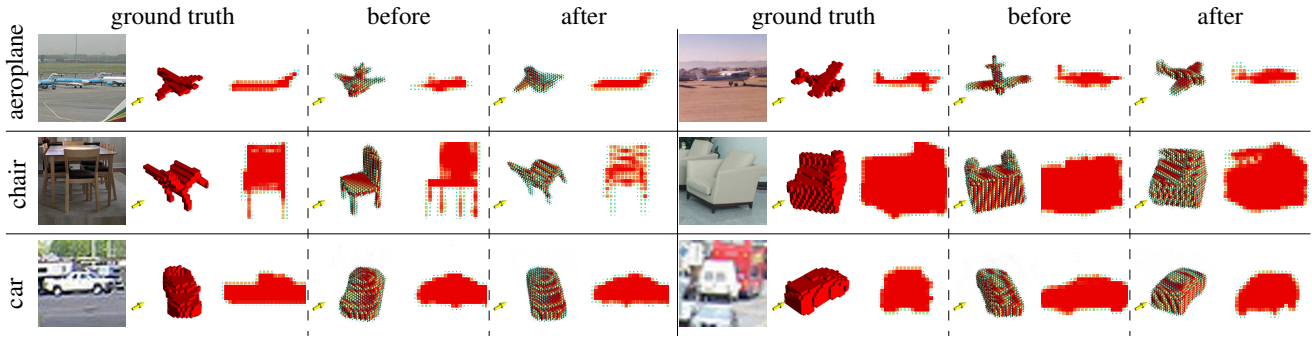

Figure 4: **Examples of improvement in pose with fine-tuning.** Identical arrangements are made as in Fig. 3 except that aligned models are not shown here.

AP) as an indication of reconstruction error. In particular, AP between aligned shapes indicates the error in style, while AP between pose-aware shapes measures the overall error in shape, influenced by both style and pose. We follow the practice in [19] in evaluation of pose estimation.

Considering the central goal of our paper is the fine-tuning scheme on natural images with reprojection loss, all these evaluations are comparisons in various metrics before and after fine-tuning. For pose estimation, we achieve comparable results as in [19]; and more importantly, we are able to further improve pose estimation with fine-tuning. For style estimation (reflected by reconstruction performance of aligned shapes in Table 4), results before fine-tuning on aligned shapes (shadowed in grey) are reflecting the performance of the prior approaches of TL-embedding network [6] and 3D-VAE-GAN [24] in our experiment setting, given this part of experiment is only re-implementation of their original works. Based on this evaluation, our approach is able to improve aligned shape reconstruction upon these two methods.

We may observe from Table 3 that, after fine-tuning by minimizing reprojection loss, not only the reprojected silhouettes better fit their ground truth, but also improvements on style (3D AP) and pose (rotation error) are achieved, in-dicating that pose-aware shapes are improved on our test set. This illustrates that the reprojection loss acts as a sufficient constraint during fine-tuning without degrading the reconstruction performance in 3D. We explain this observation with two factors. First, the generator is locked during the fine-tuning, which provides a constant prior, mapping from style parameters to shape space. As long as the fine-tuned style parameters are still within the valid scope of this generator's input space, the reconstructed shape will be valid without degradation. Second, in the fine-tuning process, the loss from rendered images in a training batch helps constrain the two regressors from over-fitting or explosion.

|  |  | p-TL | | | p-3D-VAE-GAN | | |
|---|---|---|---|---|---|---|---|
|  |  | aero | chair | car | aero | chair | car |
| 2D AP | before | 0.589 | 0.844 | 0.815 | 0.627 | 0.852 | 0.851 |
|  | after | **0.704** | **0.849** | **0.872** | **0.720** | **0.878** | **0.894** |
| 3D AP | before | 0.211 | 0.531 | 0.630 | 0.183 | 0.527 | 0.642 |
|  | after | **0.219** | **0.552** | **0.639** | **0.249** | **0.577** | **0.664** |
| rotation $Acc_{\frac{\pi}{6}}$/ $MedErr$ | before | 0.67/23.0 | 0.78/**8.2** | **0.83/4.8** | 0.67/23.2 | 0.76/8.2 | 0.86/5.0 |
|  | after | **0.68/17.3** | **0.80**/8.3 | 0.80/5.2 | **0.70/17.2** | **0.80/8.1** | 0.86/**4.7** |
|  | Su et al. | 0.76/15.1 | 0.85/9.7 | 0.86/6.1 | 0.76/15.1 | 0.85/9.7 | 0.86/6.1 |
| translation $MedErr$ | before | 0.092 | 0.074 | 0.060 | 0.088 | 0.079 | 0.061 |
|  | after | **0.077** | **0.072** | **0.058** | **0.073** | 0.079 | **0.050** |

Table 3: **Quantitative evaluation on pose-aware reconstruction.** Error in 3D & 2D shapes are measured in AP (higher is better) as in [24]. Error in rotation parameters is measured in $Acc_{\frac{\pi}{6}}$ (accuracy over $\frac{\pi}{6}$; higher is better) and $MedErr$ (median error; smaller is better) based on geodesic distance over the manifold of rotation [21]. Results form [19] are also listed. Error in translation is measured by ratio of the absolute offset against the frame size of the silhouette (30px), and we report the median number (smaller is better).

|  |  | aero | chair | car |
|---|---|---|---|---|
| p-TL | before | 0.552 | 0.709 | 0.775 |
|  | after | **0.580** | **0.731** | **0.791** |
| p-3D-VAE-GAN | before | 0.669 | 0.727 | 0.781 |
|  | after | **0.676** | **0.763** | **0.816** |

Table 4: **Quantitative evaluation on reconstruction of aligned shapes.** We also use 3D AP as the metric to measure the aligned shape error. Gray shadowed results are from re-implemented baseline methods of [6, 24].

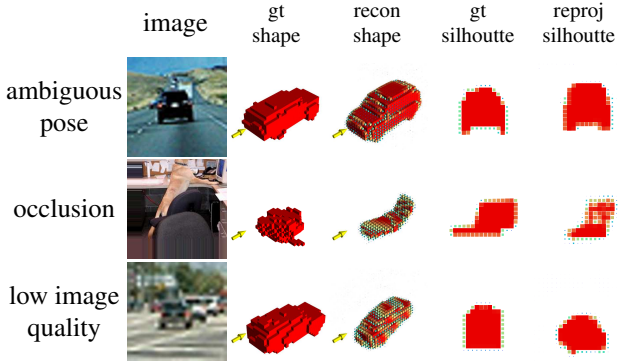

Figure 5: Failure cases in fine-tuning, mainly caused by ambiguous pose, occlusion and low image quality.

We acknowledge that in the task of shape reconstruction from a single image, optimizing the 2D reconstruction alone does not necessarily guarantee desired performance in 3D. More particularly, it can easily lead to reconstructing degraded shapes. However, by following our fine-tuning approach, which uses prior from the generator and constraint from rendered images, the reprojection loss turns out to be informative in fine-tuning as well as testing stage.

Details and analysis of fine-tuning are provided in Appendix II in supplementary material.

## 5. Conclusion

We define the new task of pose-aware shape reconstruction from a single natural image, and update the recent methods of TL-embedding Network and 3D-VAE-GAN to close the loop for this task, in both training and testing. More particularly, our proposed framework allows fine-tuning pose and style estimations by minimizing reprojection error over reprojected and ground truth silhouettes, and evaluation with this metric. The updated framework offers several advantages. First, it is capable of learning from cheaper annotation of object silhouettes in natural image sets. Second, pose and style estimations are jointly fine-tuned on natural images. Third, we demonstrate both qualitatively and quantitatively that, our fine-tuning scheme is able to not only refine pose-aware shape reconstruction, but also improve upon current state-of-the-arts on the task of aligned shape reconstruction as well as pose estimation.